\documentclass[runningheads]{llncs}
\usepackage{graphicx}
\usepackage{siunitx}
\usepackage{color}
\usepackage{amssymb}
\usepackage{amsmath, xparse}
\usepackage{makecell}
\usepackage{lipsum}
\usepackage{caption}
\usepackage{subcaption}
\usepackage{booktabs}
\usepackage{multirow}

\usepackage{pifont}
\newcommand{\bfsection}[1]{\noindent\textbf{#1.}}

\definecolor{MyBlue}{rgb}{0,0.08,0.5}
\definecolor{MyRed}{rgb}{0.7,0.02,0.02}
\definecolor{MyOrange}{rgb}{1,0.5,0}
\definecolor{MyPurple}{rgb}{0.6,0.25,0.8}
\definecolor{MyGreen}{rgb}{0.1,0.8,0.1}
\newcommand{\red}[1]{\textcolor{red}{#1}}

\newcommand{\etal}{\textit{et al}.}

\newcommand{\eg}{\textit{e}.\textit{g}.}

\usepackage[x11names]{xcolor}
\usepackage[pagebackref=false,breaklinks=true,letterpaper=true,colorlinks,bookmarks=false]{hyperref}
\hypersetup{
     colorlinks = true,
     linkcolor = red,
     anchorcolor = black,
     citecolor = SpringGreen4,
     filecolor = black,
     urlcolor = MyBlue,
     }

\newcommand*{\affaddr}[1]{#1} 
\newcommand*{\affmark}[1][*]{\textsuperscript{#1}}

\makeatletter
\def\@fnsymbol#1{\ensuremath{\ifcase#1\or \dagger\or *\or \ddagger\or
   \mathsection\or \mathparagraph\or \|\or **\or \dagger\dagger
   \or \ddagger\ddagger \else\@ctrerr\fi}}
\makeatother

\begin{document}
\title{NucMM Dataset: 3D Neuronal Nuclei Instance Segmentation at Sub-Cubic Millimeter Scale}
\titlerunning{NucMM Dataset}

\author{
Zudi Lin\affmark[1]\thanks{\footnotesize Equally contributed.} \and 
Donglai Wei\affmark[1$\dagger$] \and 
Mariela D. Petkova\affmark[1] \and
Yuelong Wu\affmark[1] \and
Zergham Ahmed\affmark[1] \and
Krishna Swaroop K\affmark[2]\thanks{\footnotesize Works were done during internship at Harvard University.} \and
Silin Zou\affmark[1] \and
Nils Wendt\affmark[3$\star$] \and
Jonathan Boulanger-Weill\affmark[1] \and
Xueying Wang\affmark[1] \and
Nagaraju Dhanyasi\affmark[1] \and
Ignacio Arganda-Carreras\affmark[4,5,6] \and
Florian Engert\affmark[1] \and
Jeff Lichtman\affmark[1] \and 
Hanspeter Pfister\affmark[1]
}

\authorrunning{Lin \etal}
\institute{
\affaddr{\affmark[1]Harvard University} \
\affaddr{\affmark[2]NIT Karnataka} \
\affaddr{\affmark[3]Technical University of Munich} \
\affaddr{\affmark[4]Donostia International Physics Center (DIPC)} \
\affaddr{\affmark[5]University of the Basque Country (UPV/EHU)} \
\affaddr{\affmark[6]Ikerbasque, Basque Foundation for Science} \\
\email{\{linzudi,donglai\}@g.harvard.edu}
}

\maketitle   
\begin{abstract}
Segmenting 3D cell nuclei from microscopy image volumes is critical for biological and clinical analysis, enabling the study of cellular expression patterns and cell lineages. However, current datasets for {\em neuronal} nuclei usually contain volumes smaller than $10^{\text{-}3}\ mm^3$ with fewer than 500 instances per volume, unable to reveal the complexity in large brain regions and restrict the investigation of neuronal structures. In this paper, we have pushed the task forward to the sub-cubic millimeter scale and curated the {\em NucMM} dataset with two fully annotated volumes: one $0.1\ mm^3$ electron microscopy (EM) volume containing nearly the entire zebrafish brain with around 170,000 nuclei; and one $0.25\ mm^3$ micro-CT (uCT) volume containing part of a mouse visual cortex with about 7,000 nuclei. With two imaging modalities and significantly increased volume size and instance numbers, we discover a great diversity of neuronal nuclei in appearance and density, introducing new challenges to the field. We also perform a statistical analysis to illustrate those challenges quantitatively. To tackle the challenges, we propose a novel hybrid-representation learning model that combines the merits of foreground mask, contour map, and signed distance transform to produce high-quality 3D masks. The benchmark comparisons on the NucMM dataset show that our proposed method significantly outperforms state-of-the-art nuclei segmentation approaches. Code and data are available at \url{https://connectomics-bazaar.github.io/proj/nucMM/index.html}.

\end{abstract}
\section{Introduction}

Segmenting cell nuclei from volumetric (3D) microscopy images is an essential task in studying biological systems, ranging from specific tissues~\cite{nhu2017novel} to entire organs~\cite{toyoshima2016accurate} and developing animals~\cite{lou2014rapid,stegmaier2016real}.
There has been a great success in benchmarking 2D and 3D nuclei segmentation methods using datasets covering samples from various species~\cite{caicedo2019nucleus,weigert2020star,tokuoka20203d,ulman2017objective}. However, existing datasets only have relatively small samples from brain tissues (\eg, volumes from mouse and rat brains~\cite{ruszczycki2019three} smaller than $10^{-3}\ mm^3$, with less than 200 instances each), restricting the investigation of neuronal nuclei in a larger and more diverse scale. Besides, most of the images are collected with optical microscopy, which can not reflect the challenges in other imaging modalities widely used in studying brain tissues, including electron microscopy (EM)~\cite{kasthuri2015saturated,hildebrand2017whole,shapson2021connectomic} and micro-CT (uCT)~\cite{dyer2017quantifying}.

\begin{figure*}[t]
    \centering
    \includegraphics[width=\textwidth]{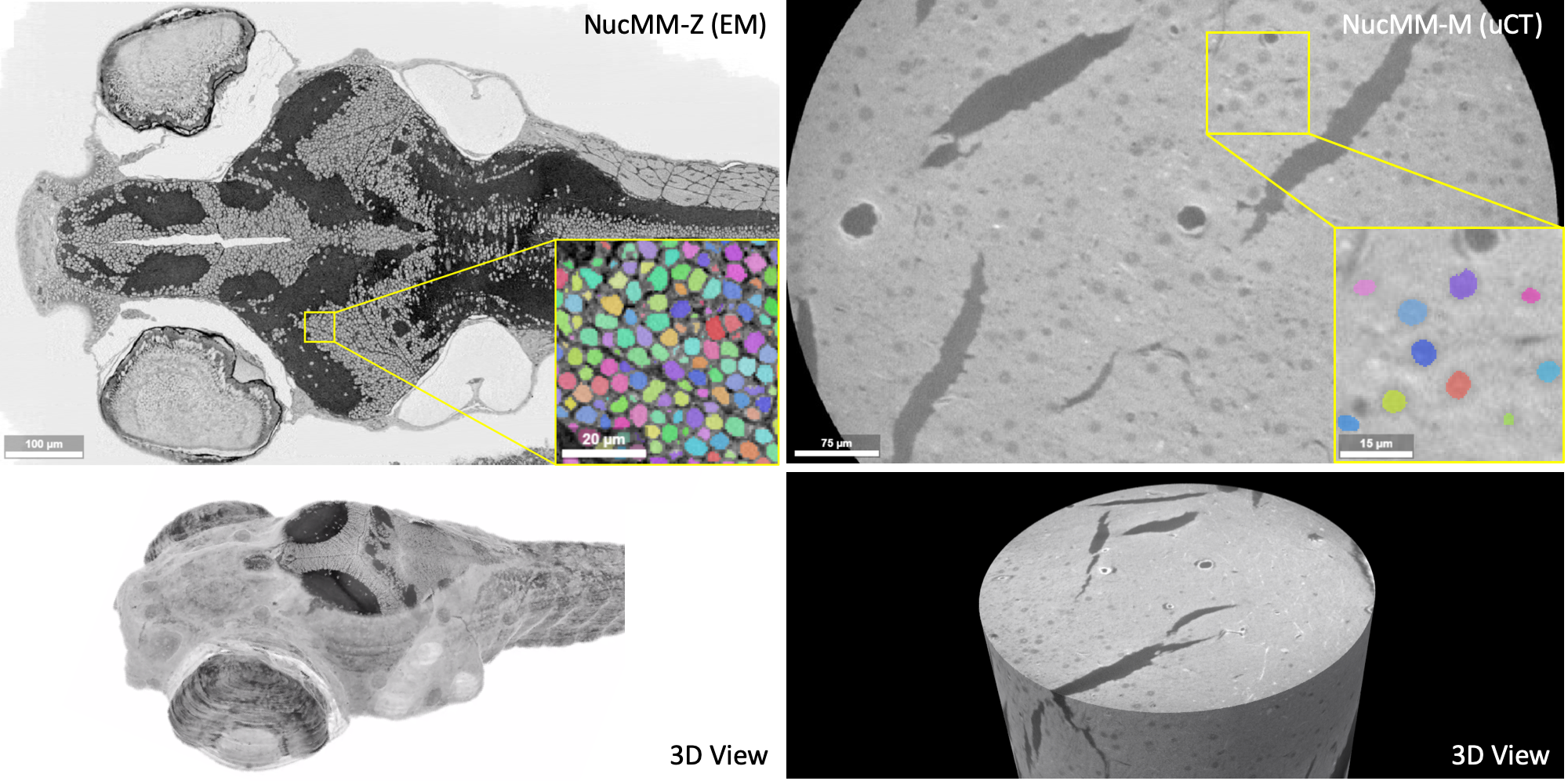}
    \caption{{\bf Overview of the NucMM dataset}. NucMM contains two large volumes for 3D nuclei instance segmentation, including {\bf (left)} the NucMM-Z electron microscopy (EM) volume 
    covering nearly a whole zebrafish brain, and {\bf (right)} the NucMM-M micro-CT volume from the visual cortex of a mouse.
    }\label{fig:teaser}
\end{figure*}

To address this deficiency in the field, we have curated a large-scale 3D nuclei instance segmentation dataset, {\bf NucMM}, which is over two magnitudes larger in terms of volume size and the number of instances than previous neuronal nuclei datasets~\cite{caicedo2019nucleus,ruszczycki2019three} and widely-used non-neuronal benchmark datasets~\cite{alwes2016live,weigert2020star}.
Our NucMM consists of one EM image volume covering a nearly entire larval zebrafish brain and a uCT image volume from the visual cortex of an adult mouse, facilitating large-scale cross-tissue and cross-modality comparison. Challenges in the two volumes for automatic approaches include the high density of closely touching instances (Fig.~\ref{fig:teaser}, left) and low contrast between object and non-object regions (Fig.~\ref{fig:teaser}, right). We also perform a statistical analysis to provide a quantitative justification of the challenges in these two volumes (Fig.~\ref{fig:stat}).

To tackle the challenges introduced by the large-scale NucMM dataset, we propose a new hybrid-representation model that learns foreground mask, instance contour map, and signed distance transform simultaneously with a 3D U-Net~\cite{cciccek20163d} architecture, which is denoted as U3D-BCD. At inference time, we utilize a multi-target watershed segmentation algorithm that combines all three predictions to separate closely touching instances and generate high-quality instance masks.
Under the average precision (AP) metric for evaluating instance segmentation approaches, we show that our U3D-BCD model significantly outperforms existing state-of-the-art approaches by $\textbf{22}\%$ on the NucMM dataset. We also perform ablation studies on the validation data to demonstrate the sensitivity of segmentation parameters.

To summarize, we have three main contributions in this paper.
First, we collected and densely annotated the NucMM dataset, which is one of the largest public neuronal nuclei instance segmentation datasets to date, covering two species with two imaging modalities.
Second, we propose a novel hybrid representation model, U3D-BCD, to produce high-quality predictions by combining the merits of different mask representations.
Third, we benchmark state-of-the-art nuclei segmentation approaches and show that our proposed model outperforms existing approaches by a large margin on the NucMM dataset. 

\subsection{Related Works}
\bfsection{Nuclei Segmentation Datasets}
Automatic cell nuclei segmentation is a long-lasting problem that is usually the first step in microscopy image analysis~\cite{meijering2012cell}. There has been a great success in benchmarking nuclei segmentation algorithms in 2D images across various types and experimental conditions~\cite{caicedo2019nucleus}. However, 2D images can not completely display the structure and distribution of nuclei. Therefore, several 3D nuclei segmentation datasets have been collected, including the Parhyale dataset showing the histone fluorescent protein expression of {\em Parhyale hawaiensis}~\cite{alwes2016live,weigert2020star}, the BBBC050 dataset showing nuclei of mouse embryonic cells~\cite{tokuoka20203d}, and multiple 3D volumes recording human cancer cells and animal embryonic cells~\cite{ulman2017objective}. However, there are few datasets covering neuronal nuclei in brain tissues, and the volumes collected from brain tissue usually contain less than 200 instances per volume~\cite{caicedo2019nucleus,ruszczycki2019three}. Besides, most public datasets mentioned above only have fluorescence images obtained with optical microscopy. 

The NucMM dataset we curated contains two volumes from animal brain tissues and covers two imaging modalities widely used in neuroscience, including EM~\cite{kasthuri2015saturated,shapson2021connectomic} and uCT~\cite{dyer2017quantifying}. In addition, our dataset is over two magnitudes larger than previous neuronal nuclei volumes~\cite{ruszczycki2019three} in terms of size and instance number. 

\bfsection{Instance Segmentation in Microscopy Images}
Instance segmentation requires assigning each pixel (voxel) not only a semantic label but also an instance index to differentiate objects that belong to the same category. The permutation-invariance of object indices makes the task challenging. For segmenting instances in microscopy images, recent learning-based approaches first train 2D or 3D convolutional neural network (CNN) models to predict an intermediate representation of the object masks such as boundary~\cite{ciresan2012deep,ronneberger2015u} or affinity maps~\cite{turaga2009maximin,lee2017superhuman}. Then techniques including watershed transform~\cite{cousty2008watershed,zlateski2015image} and graph partition~\cite{krasowski2017neuron} are applied to convert the representations into object masks. Since one representation can be vulnerable to some specific kinds of errors (\eg, small mispredictions on the affinity or boundary map can cause significant merge errors), some approaches also employ hybrid-representation learning models that learn multiple representations and combine their information in the segmentation step~\cite{stringer2021cellpose,weigert2020star,wei2020mitoem}. However, those approaches only optimize 2D CNNs~\cite{stringer2021cellpose} or learn targets that may not be suitable for nuclei segmentation~\cite{wei2020mitoem}, which leads to unsatisfactory results for down-stream analysis. Thus we propose a 3D hybrid-representation learning model that predicts foreground mask, instance contour, and signed distance to better capture nuclei with different textures and shapes.

\section{NucMM Dataset}

\bfsection{Dataset Acquisition}
The EM dataset was collected from an entire larval zebrafish brain with {\em serial-section} electron microscopy (SEM)~\cite{petkova2020correlative}. The original resolution is 30nm$\times$4nm$\times$4nm for $z$-, $y$-, $x$-axis. We downsample the images to 480nm$\times$512nm$\times$512nm (Table~\ref{tab:dataset}) to make the resolution close to other imaging modalities for nuclei analysis. The micro-CT dataset was collected from layer II/III in the primary visual cortex of an adult male mouse using 3D X-ray microscopy. The images have an isotropic voxel size of 720nm$^3$ (Table~\ref{tab:dataset}).

\begin{table}[t]
\caption{\label{tab:dataset}\textbf{NucMM dataset characteristics.} We collected and fully annotated a {\em neuronal nuclei} segmentation dataset at the sub-cubic millimeter scale. The two volumes in the dataset cover two species and two imaging modalities.}

\centering
\resizebox{\textwidth}{!}{
\begin{tabular}{l|cccccc}
\toprule
Name & Sample & Modality & Volume Size & Resolution ($\mu$m) & \#Instance\\
\midrule
NucMM-Z & Zebrafish & SEM & 397$\times$1450$\times$2000 (0.14$mm^3$) & 0.48$\times$0.51$\times$0.51 & 170K\\
NucMM-M & Mouse & Micro-CT & 700$\times$996$\times$968 (0.25$mm^3$) & 0.72$\times$0.72$\times$0.72 & 7K\\
\bottomrule
\end{tabular}
}
\end{table}

\bfsection{Dataset Annotation}
For annotating both volumes, we use a semi-automatic pipeline that first applies automatic algorithms to generate instance candidates and then asks neuroscientists to proofread the masks. Since EM images have high contrast and the object boundaries are well-defined (Fig.~\ref{fig:teaser}), we apply filtering and thresholding to get the binary foreground mask and run a watershed transform to produce the segments. For the uCT data, we iteratively enlarge the labeled set by alternating between manual correction (annotation) and automatic U-Net~\cite{ronneberger2015u} prediction. Both volumes are finally proofread by experienced neuroscientists using VAST~\cite{berger2018vast}. We also provide a binary mask for each volume to indicate the valid brain region for evaluation. 

\begin{figure*}[t]
    \centering
    \includegraphics[width=\textwidth]{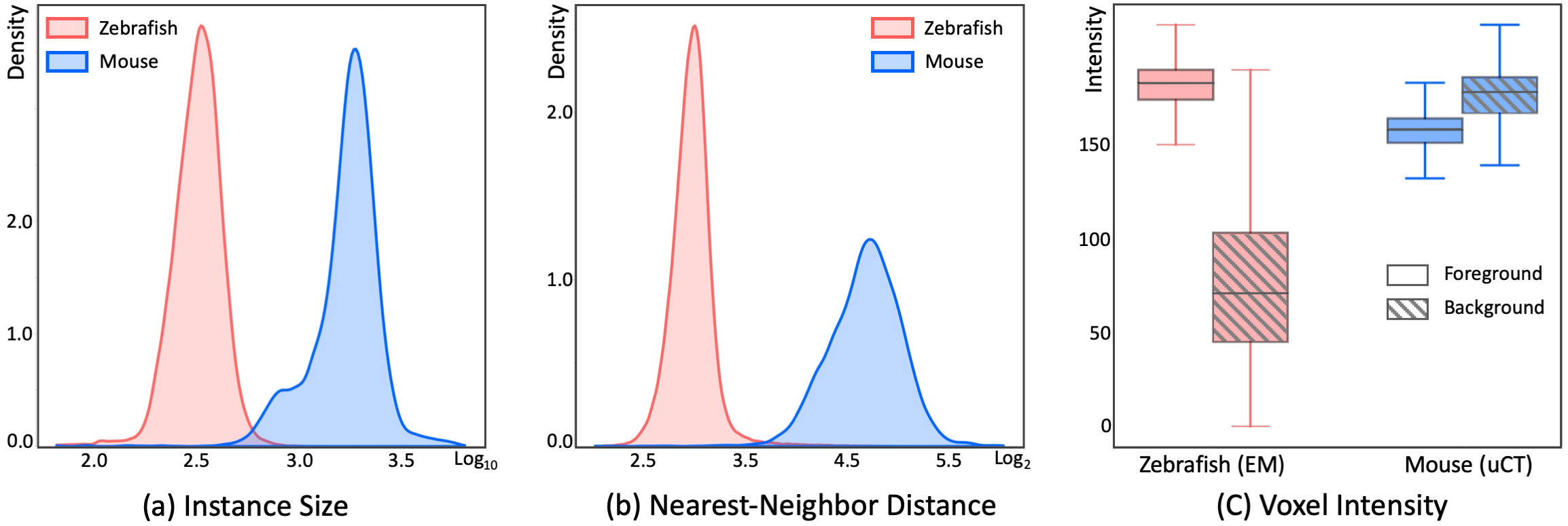}
    \caption{{\bf NucMM dataset statistics}. We show the distribution of {\bf (a)} instance size (in terms of number of voxels) and {\bf (b)} the distance between adjacent nuclei centers. The density plots are normalized by the total number of instances in each volume. We also show {\bf (c)} the voxel intensity distribution in object (foreground) and non-object (background) regions for both volumes.
    }\label{fig:stat}
\end{figure*}

\bfsection{Dataset Statistics}
The EM data has a size of 397$\times$1450$\times$2000 voxels, equivalent to a physical size of 0.14$mm^3$; the uCT data has 700$\times$996$\times$968 voxels, equivalent to 0.25$mm^3$ (Table~\ref{tab:dataset}). Although the zebrafish volume is physically smaller, it contains significantly more objects than the mouse volume, showing the distinction between brain structures. We show the distribution of nuclei size (Fig.~\ref{fig:stat}\red{a}) and the nearest-neighbor distance between nuclei centers (Fig.~\ref{fig:stat}\red{b}). The density plots illustrate that objects in the zebrafish volume are smaller and more densely packed, which poses challenges in separating closely touching nuclei.

We further show the voxel intensity distribution of object and non-object regions (Fig.~\ref{fig:stat}\red{c}). The boxplots demonstrate that objects in the EM volume are better separated from the background than the uCT volume. To quantify the separation, we calculate the {\em Kullback–Leibler} (KL) divergence between the foreground and background intensity distributions in the two volumes. The results are $D_{KL}=3.43$ and $1.33$ for the zebrafish and mouse data, respectively, showing the foreground-background contrast is significantly lower in the uCT data.

\bfsection{Dataset Splits \& Evaluation Metric}
We split each volume into $5\%$, $5\%$, and $90\%$ parts for training, validation, and testing. The limited training data makes the task more challenging, but it is also closer to the realistic annotation budget when neuroscientists handle newly collected data. Besides, to avoid sampling data in a local region without enough diversity, we follow previous practice~\cite{Januszewski2018FFN} and sample $27$ small chunks of size $64\times64\times64$ voxels from the zebrafish volume for training. Since the nuclei in the mouse uCT volume are sparser (Fig.~\ref{fig:teaser} and Fig.~\ref{fig:stat}\red{b}), we sample $4$ chunks of size $192\times192\times192$ voxels for training.

For evaluation, we use average precision (AP), a standard metric in assessing instance segmentation methods~\cite{cordts2016cityscapes,lin2014microsoft}. Specifically, we use the code optimized for large-scale 3D image volumes~\cite{wei2020mitoem} to facilitate efficient evaluation.

\section{Method}\label{sec:method}

\begin{figure*}[t]
    \centering
    \includegraphics[width=\textwidth]{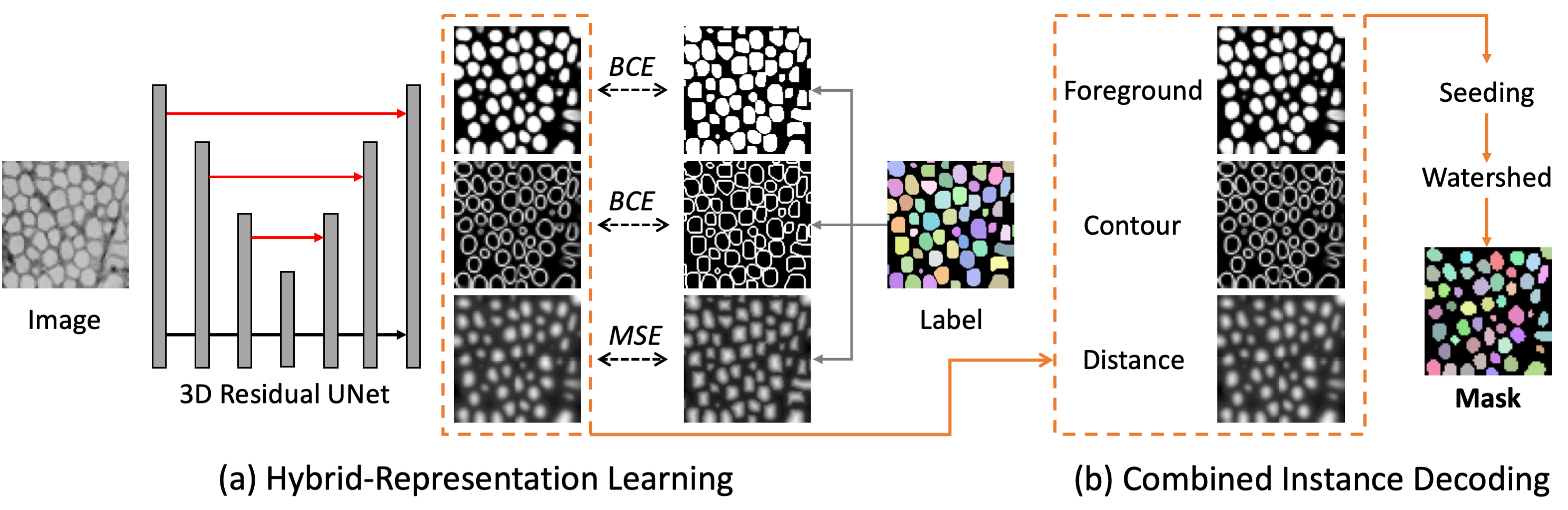}
    \caption{{\bf Hybrid-representation learning model}. {\bf (a)} Our U3D-BCD model learns a set of hybrid representations simultaneously, including foreground mask, instance contour, and signed distance transform map calculated from the segmentation. {\bf (b)} The representations are combined in seeding and watershed transform to produce high-quality segmentation masks.
    }\label{fig:method}
\end{figure*}

\subsection{Hybrid-Representation Learning}
Recent 3D instance segmentation methods for microscopy images, including Cellpose~\cite{stringer2021cellpose}, StarDist~\cite{weigert2020star} and U3D-BC~\cite{wei2020mitoem}, all use a single model to learn multiple mask representations simultaneously. Specifically, Cellpose~\cite{stringer2021cellpose} regresses the horizontal and vertical spatial gradients of the instances; StarDist~\cite{weigert2020star} learns object probability and the star-convex distance within the masks; U3D-BC~\cite{wei2020mitoem} learns the foreground mask with the instance contour map. By analyzing the representations used in those models, we notice that all the targets emphasize the learning of object masks (foreground), but the structure of the {\em background} of the segmentation map is less utilized. That is, pixels close to and far away from the object masks are treated equitably.

Therefore based on the U3D-BC model that predicts both the binary foreground mask (B) and instance contour map (C), we develop a {\bf U3D-BCD} model that in addition predicts a signed Euclidean distance map (Fig.~\ref{fig:method}\red{a}). Let $x_i$ denote a pixel in the image, we have:
\begin{equation}
  f(x_i)=\begin{cases}
    +\text{dist}(x_i, B) / \alpha, & \text{if $x\in F$}.\\
    -\text{dist}(x_i, F) / \beta, & \text{if $x\in B$}.
  \end{cases}
\end{equation}
where $F$ and $B$ denote the foreground and background masks, respectively. The scaling parameters $\alpha$ and $\beta$ are applied to control the range of the distance. Compared with the U3D-BC baseline, the signed distance map is more informative in capturing the shape information of masks. In comparison with all discussed approaches~\cite{stringer2021cellpose,weigert2020star,wei2020mitoem}, the signed distance map also model the landscape of background regions. In implementation, we apply a {\em tanh} activation to restrict the range to $(-1,1)$ and directly regress the target with a 3D U-Net~\cite{cciccek20163d}.
A similar learning target has been used for semantic segmentation of synapses~\cite{heinrich2018synaptic}, but it has not been integrated into a multi-task learning model nor has it been explored for 3D instance segmentation. Specifically, the loss we optimize is
\begin{equation}
    \mathcal{L}_{bcd} = 
    h(\sigma(y_1), y_b) + 
    h(\sigma(y_2), y_c) +
    g(\phi(y_3), y_d)
\end{equation}
where $y_i$ ($i=1,2,3$) denote the three output channels, while $\sigma$ and $\phi$ denote the sigmoid and tanh function. The foreground ($y_b$) and contour ($y_c$) maps are learned by optimizing the binary cross-entropy (BCE) loss ($h$), while the signed distance map ($y_d$) is learned with the mean-squared-error (MSE) loss ($g$).

\subsection{Instance Decoding}
In this part, we describe how to combine all three model predictions in the U3D-BCD model to generate the segmentation masks. We first threshold predictions to locate {\em seeds} (or markers) with high foreground probability and distance value but low contour probability, which points to object centers. We then apply the marker-controlled watershed transform algorithm (in the {\em scikit-image} library~\cite{scikit-image}) using the seeds and the predicted distance map to produce masks (Fig.~\ref{fig:method}\red{b}). There are two advantages over the U3D-BC~\cite{wei2020mitoem} model, which also uses marker-controlled watershed transform for decoding. First, we make use of the consistency among the three representations to locate the seeds, which is more robust than U3D-BC that uses two predictions. Furthermore, we use the smooth signed distance map in watershed decoding, which can better capture instance structure than the foreground probability map in U3D-BC. We also show the sensitivity of decoding parameters in the experiments.

\subsection{Implementation}
We use a customized 3D U-Net model that substitutes the convolutional layers at each stage with residual blocks~\cite{he2016deep}. We also change the concatenation operation to addition to save memory.
Since recent work~\cite{zhou2020towards} has shown that ADAM-alike adaptive optimization algorithms do not generalize as well as stochastic gradient descent (SGD)~\cite{bottou1991stochastic}, we optimize our models with SGD and adopt {\em cosine learning rate decay}~\cite{he2019bag}. We set the scaling parameters $\alpha$ and $\beta$ of the signed distance map to 8 and 50, respectively, without tweaking. We follow the open-source code of U3D-BC\footnote{\url{https://github.com/zudi-lin/pytorch_connectomics}} and apply data augmentations including brightness, flip, rotation, elastic transform, and missing parts augmentations to U3D-BC~\cite{wei2020mitoem} and our U3D-BCD. We also use the official implementation of StarDist\footnote{\url{https://github.com/stardist/stardist}} and Cellpose\footnote{\url{https://github.com/MouseLand/cellpose}}.

\section{Experiments}\label{sec:exp}

\begin{table}[t]
\caption{\label{tab:exp_main}
\textbf{Benchmark results on the NucMM dataset.} We compare state-of-the-art methods on the NucMM dataset using AP score. {\bf Bold} and \underline{underlined} numbers denote the 1st and 2nd scores, respectively. Our U3D-BCD model significantly improves the performance of previously state-of-the-art approaches.
}

\centering
\begin{tabular}{lccccccc}
\toprule
\multirow{2}[2]{*}{Method}
& \multicolumn{3}{c}{NucMM-Z}
& \multicolumn{3}{c}{NucMM-M}
& \multirow{2}[2]{*}{~~Overall~~}
\\\cmidrule(lr){2-4} \cmidrule(lr){5-7}
& ~AP-50~ 
& ~AP-75~ 
& ~Mean~ 
& ~AP-50~ 
& ~AP-75~ 
& ~Mean~ \\

\midrule

Cellpose~\cite{stringer2021cellpose}~ 
& 0.796 & 0.342 & 0.569 & 0.463 & 0.002 & 0.233 & 0.401\\
StarDist~\cite{weigert2020star}~ 
& \underline{0.912} & 0.328 & 0.620 & 0.306 & 0.004 & 0.155 & 0.388\\
U3D-BC~\cite{wei2020mitoem}~ 
& 0.782 & \underline{0.556} & \underline{0.670} & {\bf 0.645} & \underline{0.210} & \underline{0.428} & \underline{0.549}\\

\midrule

U3D-BCD ({\bf Ours})~
& {\bf 0.978} & {\bf 0.809} & {\bf 0.894} & \underline{0.638} & {\bf 0.250} & {\bf 0.444} &
{\bf 0.669}\\
\bottomrule
\end{tabular}
\end{table}














\subsection{Methods in Comparison}
We benchmark state-of-the-art microscopy image segmentation models including Cellpose~\cite{stringer2021cellpose}, StarDist~\cite{weigert2020star} and U3D-BC~\cite{wei2020mitoem} based on their public implementations. Specifically, the Cellpose model is trained on the 2D $xy$, $yz$, and $xz$ planes from the image volumes, and the predicted spatial gradients are averaged to generate a 3D gradient before segmentation. For StarDist, we calculate the optimal number of rays on the training data, which are 96 and 64 for the zebrafish and mouse volumes, respectively. For U3D-BC, we use the default 1.0 weight ratio between the losses of the foreground and contour map. Models are trained on a machine with four Nvidia V100 GPUs.

\subsection{Benchmark Results on the NucMM Dataset}
After choosing hyper-parameter on the validation sets, we run predictions on the $90\%$ test data in each volume and evaluate the performance. Specifically, we show the AP scores at intersection-over-union (IoU) thresholds of both 0.5 (AP-50) and 0.75 (AP-75), as well as their average (Table~\ref{tab:exp_main}). The overall score is averaged over two NucMM volumes. The results show that our U3D-BCD model significantly outperforms previous state-of-the-art models by relatively $\textbf{22}\%$ in overall performance. Besides, our method ranks 1st in 6 out of 7 scores, showing its robustness in handling different challenges. We argue that Cellpose~\cite{stringer2021cellpose} trains 2D models to estimate 3D spatial gradient, which can be ineffective for challenging 3D cases. The other two models use 3D models, but the representations StarDist~\cite{weigert2020star} uses overlooks the background in the segmentation mask, while U3D-BC~\cite{wei2020mitoem} overlooks both foreground and background structures. Although conceptually straightforward, introducing the signed distance on top of the U3D-BC baseline gives a notable performance boost.

\subsection{Sensitivity of the decoding parameters} We also show the sensitivity of the decoding hyper-parameters of our U3D-BCD model. Specifically, there are 5 thresholds for segmentation: 3 values are the thresholds of foreground probability ($\tau1$), contour probability ($\tau2$), and distance value ($\tau3$) to decide the seeds. The other 2 values are the thresholds of foreground probability ($\tau4$) and distance value ($\tau5$) to decide the valid foreground regions. 

The validation results show that the final segmentation performance is not sensitive to the $\tau1$ and $\tau2$ in deciding seeds. When fixing $\tau4=0.2$ and $\tau5=0.0$ changing the foreground probability $\tau1$ from 0.4 to 0.8 only changes the AP-50 score between 0.943 and 0.946. While changing the contour probability $\tau2$ from 0.05 to 0.30 only changes the scores from 0.937 to 0.946. However, when changing $\tau4$ within the range of 0.1 to 0.4, the score changes from 0.872 to 0.946. Those results suggest that the signed distance transform map contains important information about the object structures and performs an important role in generating high-quality segmentation masks.

\section{Conclusion}
In this paper, we introduce the large-scale NucMM dataset for 3D neuronal nuclei segmentation in two imaging modalities and analysis the challenges quantitatively.
We also propose a simple yet effective model that significantly outperforms existing approaches.
We expect the densely annotated dataset can inspire various applications beyond its original task, \eg, feature pre-training, shape analysis, and benchmarking active learning and domain adaptation methods.

\subsection*{Acknowledgments} This work has been partially supported by NSF award IIS-1835231 and NIH award U19NS104653. We thank Daniel Franco-Barranco for setting up the challenge using NucMM. M.D.P. would like to acknowledge the support of Howard Hughes Medical Institute International Predoctoral Student Research Fellowship. I.A-C would like to acknowledge the support of the Beca Leonardo a Investigadores y Creadores Culturales 2020 de la Fundación BBVA.
\bibliographystyle{splncs04.bst}
\bibliography{egbib.bib}

\end{document}